\documentclass{tufte-handout}

\usepackage[utf8]{inputenc} 

\usepackage{geometry} 
\usepackage{lscape}

\usepackage{graphicx} 
\usepackage{caption}
\usepackage[caption=false]{subfig}
\usepackage[export]{adjustbox}
\usepackage{lipsum}

\usepackage{booktabs} 
\usepackage{multirow} 

\usepackage{paralist} 
\usepackage{verbatim} 

\usepackage{array} 
\usepackage{amsmath} 
\usepackage{amsthm} 
\usepackage{amssymb} 

\usepackage{algorithm} 
\usepackage{algorithmic} 

\usepackage{tikz} 
\usepackage{pgfplots} 
\usepackage{natbib}

\makeatletter
\newcommand{\@tufte@print@margin@citation}[1]{%
    \citealp{#1}
}

\renewcommand{\@tufte@normal@cite}[2][0pt]{%
  \let\@temp@last@bibkey\@empty%
  \@for\@temp@bibkey:=#2\do{\let\@temp@last@bibkey\@temp@bibkey}%
  \sidenote[][#1]{%
    \normalsize\normalfont\@tufte@citation@font%
    \setcounter{@tufte@num@bibkeys}{0}%
    \@for\@temp@bibkeyx:=#2\do{%
      \ifthenelse{\equal{\@temp@last@bibkey}{\@temp@bibkeyx}}{%
        \ifthenelse{\equal{\value{@tufte@num@bibkeys}}{0}}{}{and\ }%
        \@tufte@trim@spaces\@temp@bibkeyx
        \@tufte@print@margin@citation{\@temp@bibkeyx}%
      }{%
        \@tufte@trim@spaces\@temp@bibkeyx
        \@tufte@print@margin@citation{\@temp@bibkeyx};\space
      }%
      \stepcounter{@tufte@num@bibkeys}%
    }%
  }%
}

\newcommand{\resetcitations}{%
  \gdef\@tufte@old@bibkeys{}%
}
\makeatother

\title[The Wilderness Area Data Set]{The Wilderness Area Data Set: Adapting the \\Covertype data set for unsupervised learning}
\author{Richard Hugh Moulton\thanks{\texttt{richard.moulton@queensu.ca}\\\noindent Department of Electrical and Computer Engineering, Queen's University, Kingston ON, Canada} and Jakub Zgraja\thanks{\texttt{jakub.zgraja@pwr.edu.pl}\\\noindent Department of Systems and Computer Networks, Wroc\l{}aw University of Science and Technology, Wroc\l{}aw, Poland}}

\begin{document}
\maketitle

\vskip\bigskipamount 
\leaders\vrule width \textwidth\vskip0.4pt 
\vskip\medskipamount 
\nointerlineskip

\begin{abstract}
Benchmark data sets are of vital importance in machine learning research, as indicated by the number of repositories that exist to make them publicly available. Although many of these are usable in the stream mining context as well, it is less obvious which data sets can be used to evaluate data stream clustering algorithms. We note that the classic Covertype data set's size makes it attractive for use in stream mining but unfortunately it is specifically designed for classification. Here we detail the process of transforming the Covertype data set into one amenable for unsupervised learning, which we call the Wilderness Area data set. Our quantitative analysis allows us to conclude that the Wilderness Area data set is more appropriate for unsupervised learning than the original Covertype data set.
\end{abstract}

\vskip\medskipamount 
\leaders\vrule width \textwidth\vskip0.4pt 
\vskip\bigskipamount 
\nointerlineskip

 \setcounter{footnote}{0} 
 
\section{Introduction}
\newthought{Benchmark data sets} are ubiquitous in the machine learning literature because they offer a method of evaluating algorithms against others in the literature as well as against a fully understood ground truth. This second requirement skews the number of benchmark data sets that are available for different tasks; the UCI Machine Learning Repository\cite{Lichman2013} contains four times more data sets intended for supervised learning than for unsupervised learning.

Even those data sets that are intended for unsupervised learning may present challenges, given the lack of an agreed upon ground truth against which to assess the eventual clusterings. Supervised learners are given labelled instances for training and can begin to infer connections between attributes and class labels. Unsupervised learners, on the other hand, are left to look at the attributes only and must discover a structure that is internal to the data set.

In this report we describe how we adapt the Covertype data set, a classic benchmark data set for supervised learning, into one that is more appropriate for use with unsupervised learning. We describe this new data set, which we call Wilderness Area, and present the results of quantitative analysis performed to confirm that the Wilderness Area data set presents a reasonable challenge for clustering algorithms.

\section{The Covertype data set}
\newthought{The Covertype dataset} was first used in a machine learning context by Blackard and Dean, as part of Blackard's doctoral thesis\cite{Blackard1998} and then as part of an academic article\cite{Blackard1999}. Both of the original works use the data set as the basis of a supervised learning task. The data set's instances are drawn from US Forest Service (USFS) Region 2 Resource Information System data and the classifier must predict the type of forest cover present in a 30 x 30 metre cell given the observed geographic information system (GIS) variables.

Since its introduction, it has become a standard benchmark data set in the literature and has been cited by hundreds of papers. The data set is available as raw data from the UCI Machine Learning Repository\cite{Lichman2013} and in normalized form from the website of Massive Online Analysis (MOA), an open source framework for data stream mining.\cite{bifet2010moa}

\subsection{Data set information}
Instances in the data set are drawn from four different wilderness areas from the Roosevelt National Forest in north Colorado: Rawah, Neota, Comanche Peak and Cache la Poudre. What makes these wilderness areas particularly useful is that they are largely the product of natural process as opposed to human management.\cite{Blackard1998} As explained in the UCI Machine Learning Repository's description of the data set\cite{Covertype1999}: 
\begin{quote}
Neota (area 2) probably has the highest mean elevational value of the 4 wilderness areas. Rawah (area 1) and Comanche Peak (area 3) would have a lower mean elevational value, while Cache la Poudre (area 4) would have the lowest mean elevational value.
\end{quote}
These instances are divided into seven types of forest cover type based on the tree species present. These are, in order, spruce/fir, lodgepole pine, Ponderosa pine, cottonwood/willow, aspen, Douglas-fir, and krummholz. Again, from the data set description\cite{Covertype1999}:
\begin{quotation}
As for primary major tree species in these areas, Neota would have spruce/fir (type 1), while Rawah and Comanche Peak would probably have lodgepole pine (type 2) as their primary species, followed by spruce/fir and aspen (type 5). Cache la Poudre would tend to have Ponderosa pine (type 3), Douglas-fir (type 6), and cottonwood/willow (type 4).

The Rawah and Comanche Peak areas would tend to be more typical of the overall data set than either the Neota or Cache la Poudre, due to their assortment of tree species and range of predictive variable values (elevation, etc.) Cache la Poudre would probably be more unique than the others, due to its relatively low elevation range and species composition.
\end{quotation}
 
\begin{figure}[htb]
\centering
\captionsetup{captionskip=0pt,farskip=0pt,nearskip=0pt}

\vspace*{-\abovecaptionskip}

\subfloat[\label{fig:covertypePCA}]{\qquad}
\includegraphics[width=0.34\textwidth,valign=T]{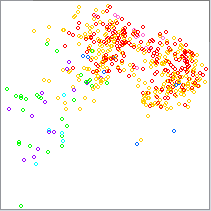}\hfil
\subfloat[\label{fig:covertypeNNMF}]{\qquad}
\includegraphics[trim=0 0 0 0.5cm,width=0.53\textwidth,valign=T]{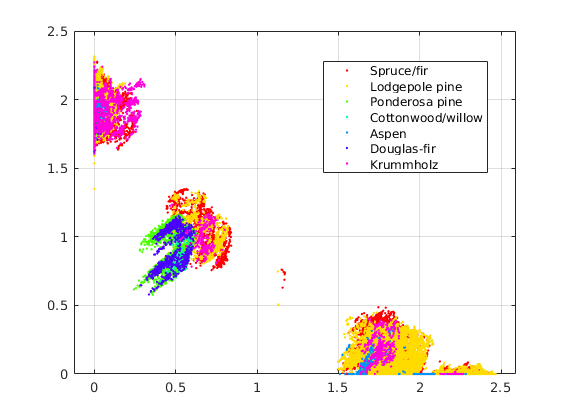}

\caption{Exploratory analysis of the Covertype dataset ---- \protect\subref{fig:covertypePCA} the first two primary components,
      \protect\subref{fig:covertypeNNMF} non-negative matrix factorization resulting in two factors.}
\label{fig:covertypeExplore}

\vspace{\abovecaptionskip}
\end{figure}

Exploratory analysis was performed using primary component analysis (PCA) in WEKA\cite{Frank2016} and non-negative matrix factorization in Matlab (Figure \ref{fig:covertypeExplore}). For both methods the data set was reduced to two dimensions to see the strongest separating effects and for ease of visualization. Inspection suggests that the forest cover type attribute does not lend itself to clusters that would be easy to learn.

\subsection{Attribute information}
The Covertype data set consists of 54 attributes that are mixed between numerical and binary attributes. An overview of these is given in Table \ref{tab:covertypeAttributes}.
\begin{table}[htb]
\begin{tabular}{p{3.25cm} c p{4.85cm}}
\toprule
Name & Data Type & Description\\
\midrule
Elevation & numeric & Elevation in metres\\[0.1cm]
Aspect & numeric & Aspect in degrees azimuth\\[0.1cm]
Slope & numeric & Slope in degrees\\[0.1cm]
Horizontal\_Distance \_To\_Hydrology & numeric & Horizontal distance to the nearest surface water features\\[0.1cm]
Vertical\_Distance \_To\_Hydrology & numeric & Vertical distance to the nearest surface water features\\[0.1cm]
Horizontal\_Distance \_To\_Roadways & numeric & Horizontal distance to the nearest roadway\\[0.1cm]
Hillshade\_9am & numeric & Hillshade index at 9 AM, summer solstice\\[0.1cm]
Hillshade\_Noon & numeric & Hillshade index at 12 PM (noon), summer solstice\\[0.1cm]
Hillshade\_3pm & numeric & Hillshade index at 3 PM, summer solstice\\[0.1cm]
Horizontal\_Distance \_To\_Fire\_Points & numeric & Horizontal distance to the nearest wildfire ignition points\\[0.1cm]
Wilderness\_Area (4) & binary & Wilderness area designation\\[0.1cm]
Soil\_Type (40) & binary & Soil type designation\\[0.1cm]
Cover\_Type & nominal & Forest cover type (7)\\
\bottomrule
\end{tabular}
\caption{The attributes contained in the Covertype data set (adapted from \emph{Blackard, 1999})}
\label{tab:covertypeAttributes}
\end{table}

Blackard gives a full description of why each attribute was included and how it was calculated\cite{Blackard1998} and we highlight two important observations here. First, elevation is an excellent predictive attribute for determining the forest cover type in a cell because most tree species in the studied wilderness areas grow within specific ranges of altitudes. This is subject to both aspect and slope, which impact both the temperature and available moisture in a given cell. Second, the 40 ``Soil Type'' attributes are very specific and could be grouped into 11 more general soil classes on the basis of USFS data.

Different subsets of attributes were tested by Blackard and Dean to ensure that each contributed information for the task of predicting forest cover type. Their results showed that classification accuracy was increased for both artificial neural networks and discriminant analysis as the number of attributes was increased.\cite{Blackard1999}

\subsection{Summary}
The Covertype data set is a very well documented data set that, due to its size, is very desirable to use as a benchmark data set for stream mining tasks. Blackard and Dean, the original authors, validated that the attributes included in the data set are useful for the cover type prediction task and the data set has been widely used in the machine learning literature.

That being said, it is clear that this data set is not conducive to clustering. In the next section we address this by transforming the data set into one that can be similarly useful for unsupervised learning.

\section{The Wilderness Area data set}
\newthought{As previous noted}, the seven forest cover types included in the Covertype dataset do not describe natural clusters. From Blackard and Dean's description, however, we note that the Wilderness Area attributes have the kind of semantic meaning that we would expect to be able to learn and represent well using a clustering algorithm.\cite{Blackard1999}

We therefore make use of the normalized data set available from the MOA website and switch the class label in order to support a change in task. Now instead of predicting the forest cover type, the objective for the learner is now to cluster instances according to their ground truth Wilderness Area.

\subsection{Exploratory analysis}
\begin{figure}[htb]
\centering
\captionsetup{captionskip=0pt,farskip=0pt,nearskip=0pt}

\vspace*{-\abovecaptionskip}

\subfloat[\label{fig:wildernessPCA}]{\qquad}
\includegraphics[width=0.35\textwidth,valign=T]{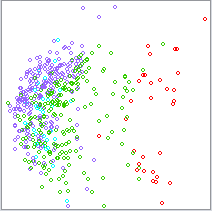}\hfil
\subfloat[\label{fig:wildernessNNMF}]{\qquad}
\includegraphics[trim=0 0 0 1cm,width=0.5\textwidth,valign=T]{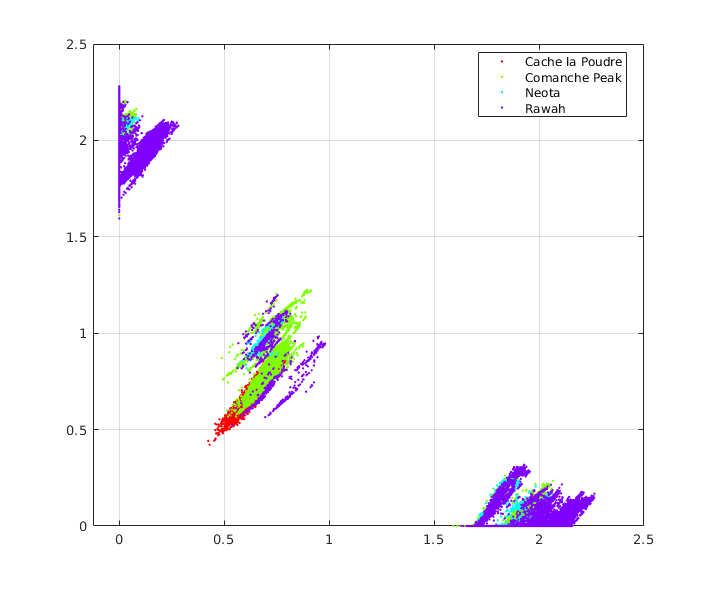}

\caption{Exploratory analysis of the Wilderness Area dataset ---- \protect\subref{fig:wildernessPCA} shows the first two primary components,
      \protect\subref{fig:wildernessNNMF} shows non-negative matrix factorization resulting in two factors.}
\label{fig:wildernessExplore}

\vspace{\abovecaptionskip}
\end{figure}

We begin by performing the same exploratory analysis as was done for the Covertype data set: PCA using WEKA and non-negative matrix factorization using MATLAB. The results are shown in Figure \ref{fig:wildernessExplore} and appear to indicate a higher degree of separation than was seen with the Cover\_Type attribute.

Very noticeable is that the Cache la Poudre wilderness area is clearly the most distinct of the four, as assessed in the description for the Covertype data set.\cite{Covertype1999}

\subsection{Dimensionality}
One aspect of the Covertype data set that made it very difficult for clustering algorithms is its dimensionality. This leads to the curse of dimensionality, which has negative effects on distance metrics and the clusters that are based on them 

To address this issue we make an effort to faithfully represent the data set in as low a dimension as possible. We do this by merging the 40 binary Soil\_Type attributes into one nominal Soil\_Type attribute and by keeping the Cover\_Type attribute as a single nominal attribute rather than splitting it into seven binary attributes as was done for the Wilderness Area attributes in the Covertype data set.

The result of this processing is a twelve-dimensional vector of attributes and a single class attribute. This is fewer than a quarter of the attributes for the Covertype data set without sacrificing any of the GIS data represented. The attributes for the Wilderness Area data set correspond to the rows in Table \ref{tab:covertypeAttributes} with nominal attributes replacing binary.

\section{Quantitative Analysis}
\newthought{Although we might be satisfied} with our intuitions as laid out in the previous section, we have also conducted quantitative analyses to ensure that we have achieved our goal.

\subsection{Attributes}
At the attribute level, we calculate each attribute's Pearson correlation coefficient and Information Gain with respect to its data set's label using Weka.\cite{Frank2016} Figure \ref{fig:attributeMeasures} shows both of these measures for the top 12 attributes from the two data sets.
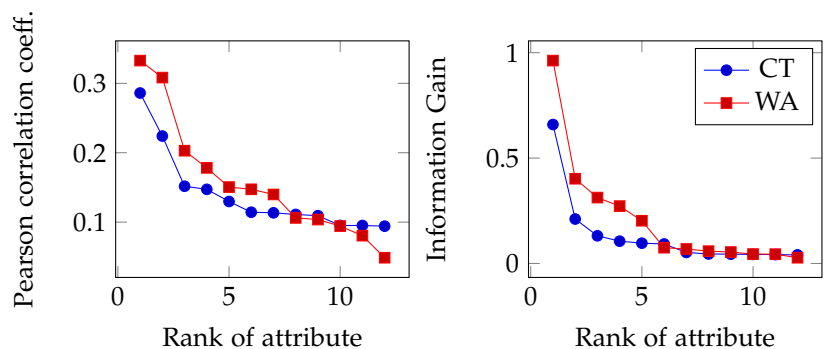
\begin{figure} \label{fig:attributeMeasures}
\centering
\begin{tikzpicture}
      \begin{axis}[
          width=0.5\linewidth, 
          xlabel=Rank of attribute, 
          ylabel=Pearson correlation coeff.,
        ]
        \addplot
        table[x=attribute,y=covertype,col sep=comma] {correlation.csv}; 
        \addplot
        table[x=attribute,y=wilderness,col sep=comma] {correlation.csv}; 
      \end{axis}
    \end{tikzpicture}
\begin{tikzpicture}
      \begin{axis}[
          width=0.5\linewidth, 
          xlabel=Rank of attribute, 
          ylabel=Information Gain,
          legend pos=north east,
        ]
        \legend{CT,WA}
        \addplot
        table[x=attribute,y=covertype,col sep=comma] {infoGain.csv}; 
        \addplot
        table[x=attribute,y=wilderness,col sep=comma] {infoGain.csv}; 
      \end{axis}
    \end{tikzpicture}
\caption{The utility of different attributes in predicting the class label for the Covertype and Wilderness Area data sets.}
\label{fig:attributeMeasures}
\end{figure}

It is clear from both graphs that transforming the data set has had a positive effect on how well each individual attribute relates to the ground truth labelling: the highest ranking attributes for both measures score higher for the Wilderness Area data set while the remaining attributes generally show fairly even scores.

\subsection{Clusters}
At the cluster level, we measure the silhouette coefficient for both data sets assuming a ``perfect'' clustering where the ground truth labels are used to indicate cluster membership The silhouette coefficient for a clustering is the mean silhouette value across all instances in the data set and it ranges between $-1$, which indicates the poorest clustering, and $1$, which indicates the best clustering.

The silhouette coefficient is useful because it is an internal measure of cluster quality, meaning we can use it to assess how good the clusters represented by a given labelling are. It was used by Kremer et al. as a benchmark measure for their design of a new external measure of cluster quality.\cite{Kremer2011}

\begin{figure*}[htb]
\centering
\captionsetup{captionskip=0pt,farskip=0pt,nearskip=0pt}

\vspace*{-\abovecaptionskip}

\subfloat[\label{fig:covertypeSilhouette}]{\qquad}
\includegraphics[width=0.45\textwidth,valign=T]{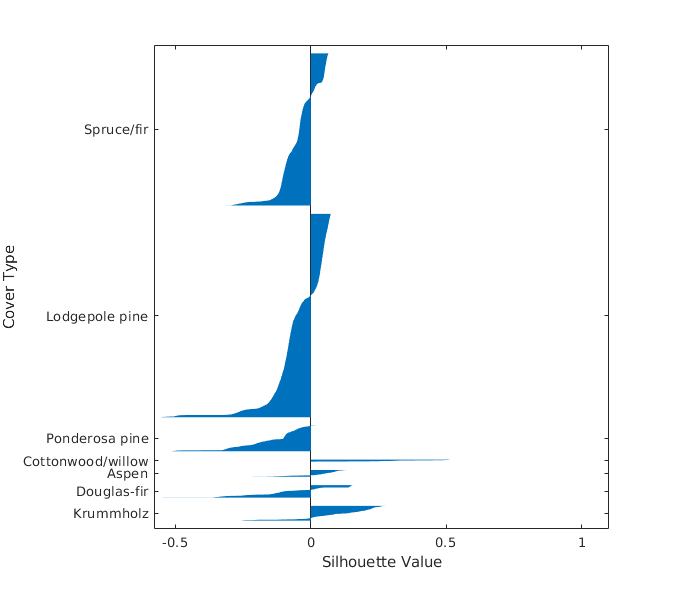}\hfil
\subfloat[\label{fig:wildernessSilhouette}]{\qquad}
\includegraphics[width=0.45\textwidth,valign=T]{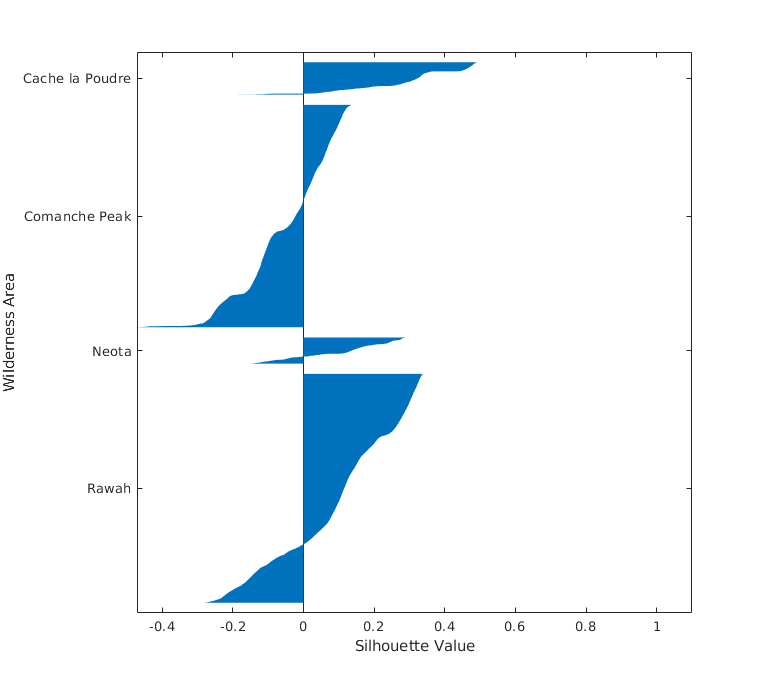}

\caption{Silhouete analysis ---- \protect\subref{fig:covertypeSilhouette} the Covertype data set; and
      \protect\subref{fig:wildernessSilhouette} the Wilderness Area data set.}
\label{fig:silhouetteAnalysis}

\vspace{\abovecaptionskip}
\end{figure*}

MATLAB was used to calculate the silhouette coefficients for both data sets. For the purposes of computational feasibility, the silhouette coefficient was calculated for the first of ten stratified folds for both of the data sets. These were $-0.045$ for Covertype and $0.0477$ for Wilderness Area (Figure \ref{fig:silhouetteAnalysis}). While the difference in values is small, the silhouette coefficient for the Wilderness Area data set is higher and is above $0$ -- both of which indicate a more natural clustering.

\section{Conclusion}
\newthought{In this report} we have detailed the Covertype data set and the reasons why it has become a classic benchmark data set for supervised learning tasks in the machine learning literature. We also noted, however, that it was not well adapted to being used as a benchmark data set for unsupervised learning.

Inspired by the thorough documentation of the data set, we therefore transformed the Covertype data set into the Wilderness Area data set. Using the same domain and semantic meaning, the Wilderness Area data set changes the task from predicting the forest cover type to clustering instances by wilderness area. The quantitative analysis that we performed confirms that, although the clustering task will remain challenging, the Wilderness Area data set is more conducive to finding clusters than the original Covertype data set.

\bibliography{WildernessArea.bib}
\bibliographystyle{plainnat}

\end{document}